%% file: main.tex
\documentclass[sigconf,nonacm]{acmart}
\usepackage{enumitem}
\usepackage{stfloats}
\usepackage{multirow}
\usepackage{tikz}
\usetikzlibrary{shapes,arrows.meta,positioning}

\newcommand{\bfhead}[1]{\vspace{0.2cm}\noindent\textbf{#1}}

\newcommand{\vtext}[2]{\parbox[t]{2mm}{\multirow{#1}{*}{\rotatebox[origin=c]{90}{#2}}}}
\newcommand{\cm}{\checkmark}%
\newcommand{\tval}[3]{, $t(#1)=#2$, $p=#3$}
\begin{document}

\title{Studying the Effects of Robot Intervention \\on School Shooters in Virtual Reality}

\author{Christopher A. McClurg}
\orcid{0000-0002-0558-4637}
\affiliation{%
  \institution{The Pennsylvania State University}
  \department{Dept. of Aerospace Engineering}
  \city{State College}
  \state{PA}
  \country{USA}
}
\email{cam7498@psu.edu}

\author{Alan R. Wagner}
\affiliation{%
  \institution{The Pennsylvania State University}
  \department{Dept. of Aerospace Engineering}
  \city{State College}
  \state{PA}
  \country{USA}
}
\email{azw78@psu.edu}

\renewcommand{\shortauthors}{McClurg and Wagner}

 % -------------------------------------------------------------------
\begin{abstract}
    We advance the understanding of robotic intervention in high-risk scenarios by examining their potential to distract and impede a school shooter. To evaluate this concept, we conducted a virtual reality study with 150 university participants role-playing as a school shooter. Within the simulation, an autonomous robot predicted the shooter’s movements and positioned itself strategically to interfere and distract. The strategy the robot used to approach the shooter was manipulated -- either moving directly in front of the shooter (aggressive) or maintaining distance (passive) -- and the distraction method, ranging from no additional cues (low), to siren and lights (medium), to siren, lights, and smoke to impair visibility (high). An aggressive, high-distraction robot reduced the number of victims by 46.6\% relative to a no-robot control. This outcome underscores both the potential of robotic intervention to enhance safety and the pressing ethical questions surrounding their use in school environments. 
\end{abstract}

 % -------------------------------------------------------------------
%% http://dl.acm.org/ccs.cfm.

%\ccsdesc[500]{Human-centered computing~Empirical studies in interaction design}
%\ccsdesc[500]{Human-centered computing~Virtual reality}
%\ccsdesc[100]{Hardware~Safety critical systems}
 % -------------------------------------------------------------------

 % -------------------------------------------------------------------
%\received{26 August 2025}

 % -------------------------------------------------------------------
\maketitle

 % -------------------------------------------------------------------
\section{Introduction}
Regrettably, school shootings in the United States have increased dramatically. According to the K--12 School Shooting Database, there were more school shootings from 2021--2024 (1244) than in the years 1970-2016 (1233) \cite{riedman2018k}. The harm caused by school shootings goes beyond the immediate victims. Students who witness a shooting suffer high rates of Post-Traumatic Stress Disorder (PTSD) and psychiatric disorders \cite{elklit2013psychological, suomalainen2011controlled}. 

Various approaches have been implemented to prevent school shootings. These include hardening school buildings, adding security measures such as metal detectors, zero tolerance bullying programs, and profiling potential active shooters \cite{mongan2009etiology}. The effectiveness of these efforts is unclear \cite{jonson2017preventing, crawford2015preventing, borum2010can}. More than half of schools have hired armed school resource officers~\cite{jonson2017preventing, addington2009cops, us2015results}, even though armed officers are a substantial cost \cite{addington2009cops}, may signal to students that schools are unsafe \cite{bachman2011predicting}, and may not be an effective deterrent~\cite{james2013school}.       

\begin{figure}[t!]
  \centering
  \includegraphics[width=\linewidth]{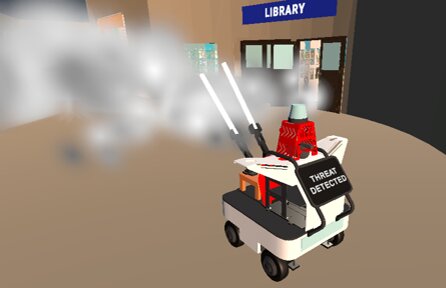}
  \caption{This study pilots a response robot that distracts a shooter with loud noises, flashing lights, and/or smoke.}
  \label{fig:robot_with_smoke}
\end{figure}

This paper examines the possibility of using a robot (see Figure~\ref{fig:robot_with_smoke}) to intervene in a school shooting to delay and/or distract a school shooter, slowing their progress and potentially saving lives. Rather than having a dedicated active shooter intervention robot, we envision that robots already present in the school for some other purpose, such as cleaning, could be activated during an emergency by school alarms or local emergency personnel. In response to being activated, the robot would (1) locate the shooter using school surveillance cameras, (2) predict the future location of the shooter, (3) use the predicted shooter location to approach the shooter, and (4) act to distract the shooter. This paper focuses on the last two steps of that response. The overarching goal is to delay the shooter as long as possible, ideally until the arrival of law enforcement.

We assume that the robot is notified of the shooter's presence and relies on a closed-circuit camera system to track the real-time location of students and the shooter. Prior work has successfully developed methods for tracking people during a simulated emergency \cite{nayyar2023modeling, nayyar2023near}. We also assume that students and faculty will shelter in place, as recommended and documented in previous school shootings \cite{ausdemore2015eliminating}. We presume that the school has been mapped and can be remapped nightly, allowing the robot to quickly and accurately localize using on-board sensors (IMU, lidar, camera) and standard algorithms from the Robot Operating System (ROS) framework \cite{quigley2009ros}. Given these assumptions, the robot can navigate to the general location of the shooter.

As a step towards developing a robot that can intervene during school shootings, we conducted a human-subject study in virtual reality in which university students and faculty were asked to act as if they were a school shooter during a simulated shooting. We used this study to compare two different strategies for approaching the shooter and three different methods for distracting and delaying the shooter. This paper contributes evidence that robots could be capable of delaying an active shooter and, thus, reducing the number of victims. To the best of our knowledge, this is the first work to explore the use of robots to intervene during a school shooting.

% -------------------------------------------------------------------
\section{Related Work}
In this application, the human shooter and the robot are adversaries. Therefore, we present related work in the areas of adversarial robotics and human compliance with robot commands. These areas are described below. 

\subsection{Adversarial Robotics}
Non-lethal adversarial robots have been developed for law enforcement and patrol applications~\cite{nguyen2001robotics, lundberg2007assessment, foxnewsChinaRolls,cramer2021digidog}. A survey of law enforcement agencies indicated that, if available, robots would be used frequently to inspect hazardous areas or as a tool for hostage negotiation~\cite{nguyen2001robotics}. In another study, a special weapons and tactics unit (SWAT) was given a mobile robot with camera and siren for five months. By the end of this period, they concluded that the robot was suitable for hostage negotiation~\cite{lundberg2007assessment}. Law enforcement in China has introduced a 275-pound spherical robot, capable of speeds up to 22 miles per hour, to track and pursue suspects~\cite{foxnewsChinaRolls}. In the United States, law enforcement has tested a four-legged robot dog for similar purposes, where it has shown promise in hostage situations~\cite{cramer2021digidog}.

Adversarial multi-robot patrolling is another application related to this work~\cite{huang2019survey}. Whereas regular patrolling is focused on area coverage and threat detection, adversarial patrolling assumes the presence of an intelligent adversary~\cite{huang2019survey}. Agmon et al. introduced an algorithm for multi-robot patrolling that uses a random Markovian strategy to improve robustness against adversaries that have learned the patrol strategy~\cite{agmon2011multi}. Sless et al. present an algorithm for multi-robot patrolling that considers multiple coordinated attacks~\cite{sless2014multi}. Alam et al. provide a decentralized method for multi-robot patrolling for situations where communication is limited ~\cite{alam2015distributed}. Their results indicate that the method offers performance comparable to centralized baselines. Finally, Talmor and Agmon demonstrate that a patrolling robot team could use deception to make an environment appear more secure, and hence use resources more effectively~\cite{talmor2017power}. 

\subsection{Human Compliance in HRI} 
Researchers have also considered how robot-related factors contribute to a person's willingness to comply with a robot’s request or command~\cite{cormier2013would, agrawal2017robot, herzog2022influence, hou2023should}. With respect to \textit{robot embodiment}, Agrawal and Williams showed that people more often comply with a robot acting as security guard when they perceived it to be more human-like~\cite{agrawal2017robot}. Jois and Wagner, on the other hand, found that people were less likely to comply with robot-administered punishments and may not accept a robot's authority to punish \cite{jois2021happens}. Cormier et al. found that participants who are asked to complete a tedious computer task protested significantly more to a robot than a human running the experiment~\cite{cormier2013would}. Other studies varied the robot's anthropomorphism but found no significant influence on compliance~\cite{geiskkovitch2015autonomy, geiskkovitch2016please, haring2019robot, herzog2022influence}. With respect to \textit{robot speech}, no significant differences were observed by simply varying the tone (playful versus factual) as a tool to persuade participants~\cite{8956298, kharub2022effectiveness}. However, Goetz et al. found that compliance increased when the robot tone matched the seriousness of the task~\cite{1251796}. Other research has shown that people comply more when a robot uses negative language instead of positive~\cite{midden2009using}, coercive language instead of offering reward~\cite{9223608}, social language rather than functional~\cite{horstmann2018robot}, or emotional language rather than logical~\cite{9093955}. With respect to \textit{social attributes}, the robot's credibility was not found to impact compliance~\cite{lee2016role}. Siegel et al. examined the effect of perceived gender, showing that men complied significantly more often with the female-sounding robot, while perceived gender had no effect on female participants~\cite{ siegel2009persuasive}. With respect to \textit{robot authority}, the results have been mixed. Sembroski et al. demonstrated that participants were more likely to comply with a robot perceived to have greater authority than the experimenter~\cite{8172280}. In contrast, Saunderson and Nejat showed that participants were more likely to comply with a non-authoritative robot~\cite{saunderson2021}. Finally, Hou et al. showed that when a participant was teamed with a human and a robot, he/she was significantly more influenced by the agent with authority, whether human or robot~\cite{ hou2023should}. 

% -------------------------------------------------------------------
\section{Experimental Setup}
\begin{figure*}[t!]
  \centering
  \includegraphics[width=0.8\linewidth]{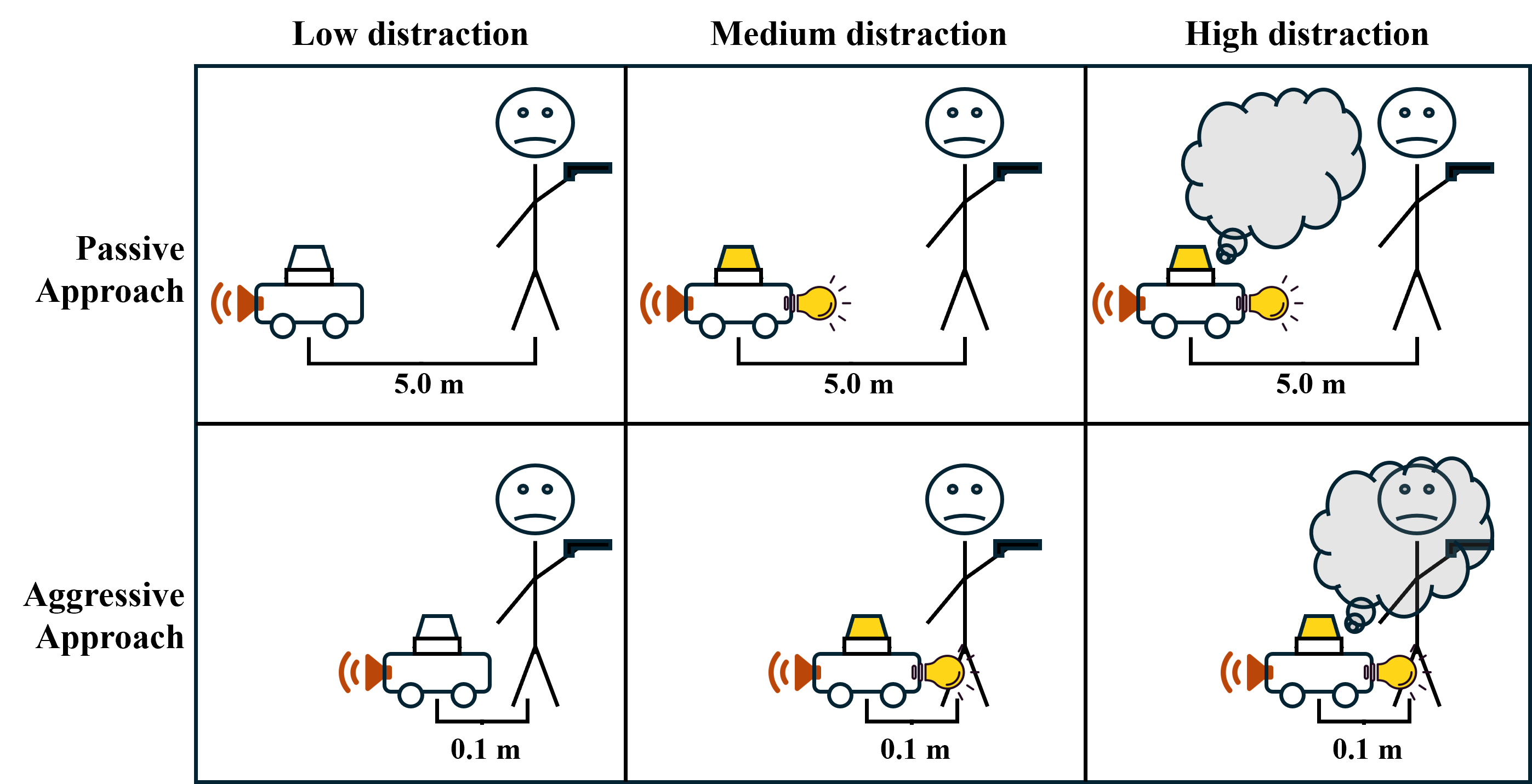}
  \caption{The different study conditions are depicted. These conditions were compared to a ``no robot" control condition.}
  \label{fig:exp_design}
\end{figure*}

We conducted an experiment in virtual reality that required participants to play the role of a school shooter. The experiment was conducted in a virtual replication of Columbine High School, the site of an infamous school shooting in the United States. The environment was populated with non-player characters (NPCs) acting as students and teachers. For five minutes, the participant tried to shoot as many NPCs as possible, while two robots (one per floor) tried to intervene. As an initial investigation of the potential of robots to intervene and distract school shooters, we manipulated two independent variables. First, we consider how the robot approaches a shooter. The approach style was either \textit{passive}, where the robot followed the participant by five meters, or \textit{aggressive}, where the robot raced to the shooter's predicted future position (five seconds ahead). Second, the method used by the robot to distract was manipulated. Three types of distractions were used. As a baseline, the mere presence of the robot itself was the only distractor. Next, we considered the use of distracting lights (strobes) and sounds (a siren) on the robot. Finally, the highest level of distraction involved lights, sirens, and smoke. These conditions are referred to by their level of distraction, i.e., \textit{low}, \textit{medium}, and \textit{high}. Figure~\ref{fig:exp_design} depicts the 2x3 experimental design that was used. These experimental conditions were compared to a ``no robot'' control condition from a previous study~\cite{mcclurg2025using}. 

\bfhead{Participants.} University students and faculty were recruited using poster and email advertisements. Because 96-97\% of school shooters are male, the inclusion criteria for the study required participants to be male~\cite{peterson2023violence}. The inclusion criteria also required subjects to be at least 18 years old and not prone to motion sickness. Subjects were initially paid \$15 per hour but this rate was later increased to \$30 per hour to attract more participants. Individuals were not allowed to participate more than once. This study was IRB approved.

\begin{figure}[!b]
  \centering
  \includegraphics[width=\linewidth]{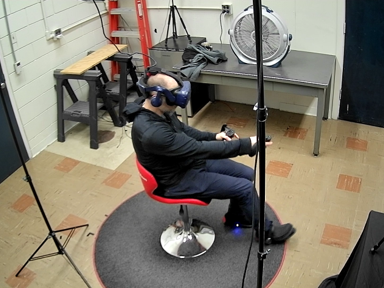}
  \caption{Participants were seated in a swivel chair with an HTC Vive headset. Shoe interfaces (Cybershoes) were strapped over their shoes allowing them to move through the simulation by making a walk motion. Valve Index controllers were strapped to hands for hand movements and shooting.}
  \label{fig:participant}
\end{figure}

\bfhead{Dependent Variables.} The dependent variables included participant distance traveled, the number of victims, the number of shots fired, the portion of shots fired at the robot, and the portion of time the participant looked at the robot (gaze). Among these, we focus on the number of victims ($V$) and gaze directed at the robot ($G^R$) because these variables inform us about the potential value of the research and why our approach might work. The participant's position, rotation, gaze direction, eye pupil diameter, number of shots fired and reloads, as well as the position and visibility of objects were recorded at a rate of 2 Hz. The data related to these variables is available on GitHub (link redacted). %\href{https://github.com/chrismcclurg/VR-Shooter-Data}{GitHub}.

\begin{figure*}[!t]
  \centering
  \includegraphics[width=\linewidth]{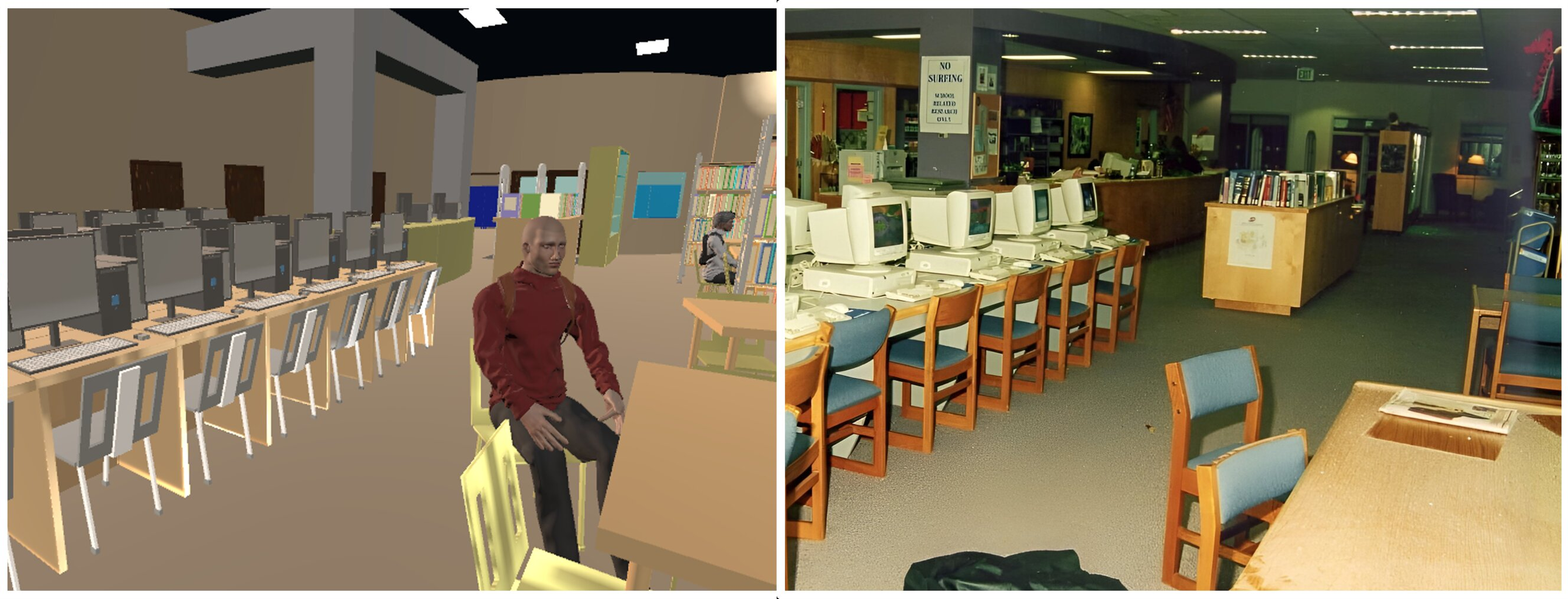}
  \caption{The simulation environment (left) was a replica of Columbine High School (right). A finite-state machine was used to control the NPCs behavior. The visible NPCs on the left are in a relaxed state. The real image on the right was released by Bill Ockham in 2023~\cite{klebold_trenchcoat_2023}.}
  \label{fig:real_vs_sim_library}
\end{figure*}

\bfhead{Hypotheses.} 
A total of 45 hypotheses were pre-registered on the Open Science Framework (OSF), many of which were exploratory. For this paper, we focus on a subset\footnote{Only reporting a subset of hypotheses did not affect our usage of statistical methods for multiple comparison correction~\cite{holm1979simple}. We still used the total number of hypotheses within each family of hypotheses to determine the appropriate adjustment.} of hypotheses related to our selected dependent variables ($V$, $G^R$). The full set of hypotheses are available on OSF (link redacted).
%\href{https://osf.io/angy3/?view_only=7a1f71c48b2a44669efd87f307bf78ce}{OSF}. 

\vspace{0.4cm}
\begin{enumerate}[label=H\arabic*.]
\setlength{\itemsep}{4pt}
\item The number of victims ($V$) will decrease with increasingly aggressive robot approach.
\item The number of victims ($V$) will decrease with increased level of distraction.
\item The shooter's attention to the robot ($G^R$) will increase with increasingly aggressive robot approach. 
\item The shooter's attention to the robot ($G^R$) will increase with increased distraction level.
\item The presence of a robot will reduce the number of victims ($V$) relative to the baseline condition without a robot.
\end{enumerate}

\input{figures/procedure}
\vspace{1cm}
\bfhead{Procedure.} The procedure is depicted in Figure~\ref{fig:procedure}. Upon arrival to the lab, the participant was asked to complete a consent form. The participant was then outfitted with virtual reality equipment, enabling them to view a virtual environment that resembled the lab where the experiment was taking place. An NPC resembling the experimenter then led the participant first through a training procedure which allowed them to practice moving and shooting in the virtual environment. Next, the NPC led participants on a tour of the school highlighting points of interest. Because most school shooters have detailed knowledge of the school, the purpose of this tour was to increase the subject's familiarity with the school. The participant was then given a handgun. Finally, NPCs representing students, faculty, and staff arrived on campus. The shooting began once the participant fired their first shot and continued for a total of five minutes. In all test conditions, it was assumed that surveillance cameras could provide the robot with the shooter's current location, which then served as input to our shooter position model to predict the future position of the shooter. %Figure~\ref{fig:procedure} depicts the procedure used for this experiment. 

\begin{figure}[b!]
  \centering
  \includegraphics[width=0.6\linewidth]{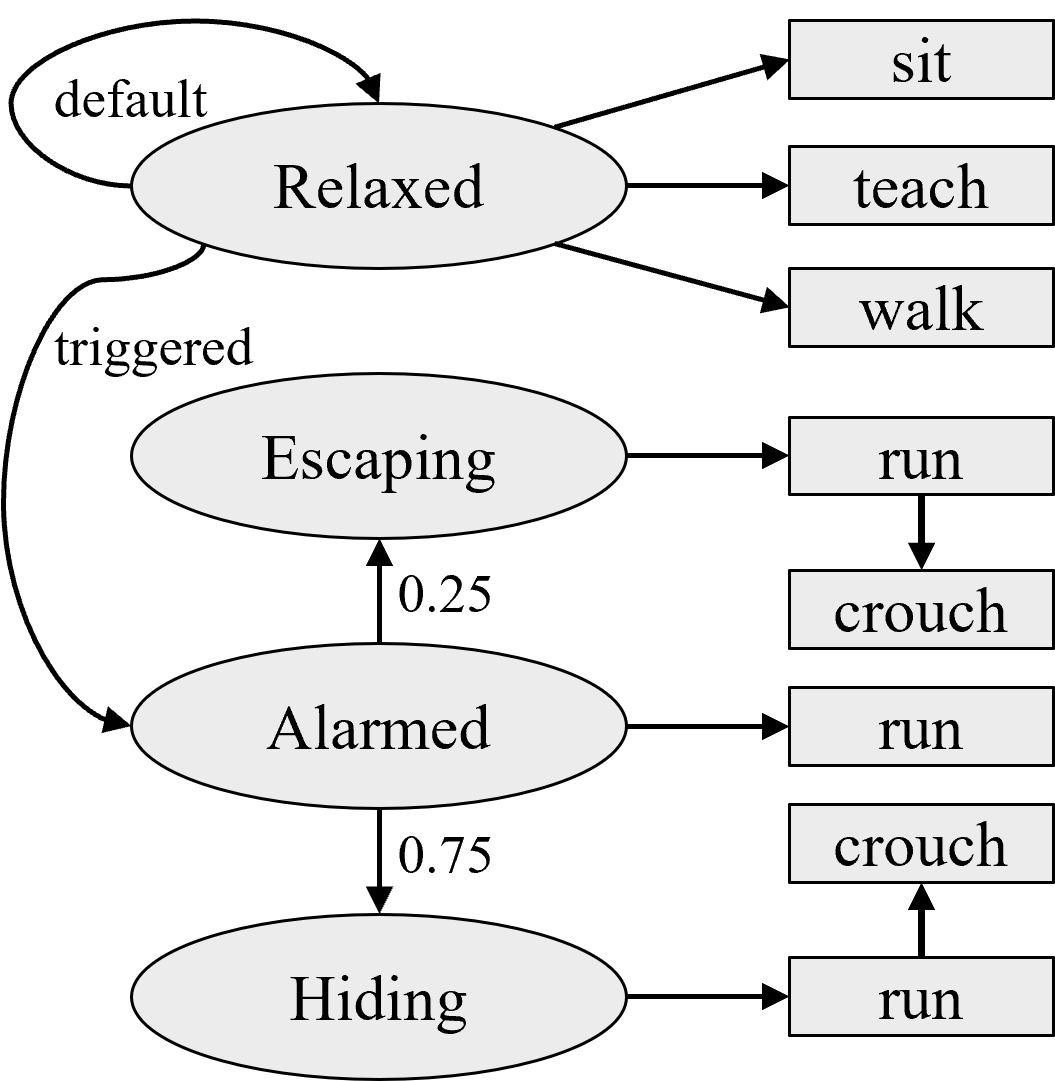}
  \caption{A finite state machine is used to guide the behavior of each NPC. The column of ovals describes an NPCs possible behavioral states for the experiment. The lines denote transitions between different states. The numbers associated with a transition represent the probability of the transition. NPCs remain in a 'Relaxed' state until they are triggered by seeing the shooter or alarmed NPC. The column of rectangles describes the animation used to generate each behavior in the simulation. }
  \label{fig:fsm}
\end{figure}

\bfhead{Non-Player Characters (NPCs).} Non-player characters (NPCs) were controlled by a finite state machine (FSM), a common practice for modeling evacuees~\cite{jayaparvathy2021study, che2015novel, kielar2014concurrent}. The FSM was designed to mimic the ``Run. Hide. Fight'' response recommended by the Department of Education ~\cite{us2013guide}. Fighting is difficult to accurately simulate and rarely occurs during actual school shootings, so this option was disregarded. The resultant FSM is shown in Figure~\ref{fig:fsm}. The default state of the NPC was relaxed. When triggered by seeing the shooter, victim, or alarmed individual, the NPC would transition to an alarmed state. From the alarmed state, the NPC would transition to a hiding state with a 75\% probability or to an escaping state with 25\% probability. %the hiding or escaping states randomly. Our model includes the run-hide decision probability as a parameter, which we set to have the NPCs attempt to escape 25 percent of the time. 

\begin{figure}[h!]
  \centering
  \includegraphics[width=\linewidth]{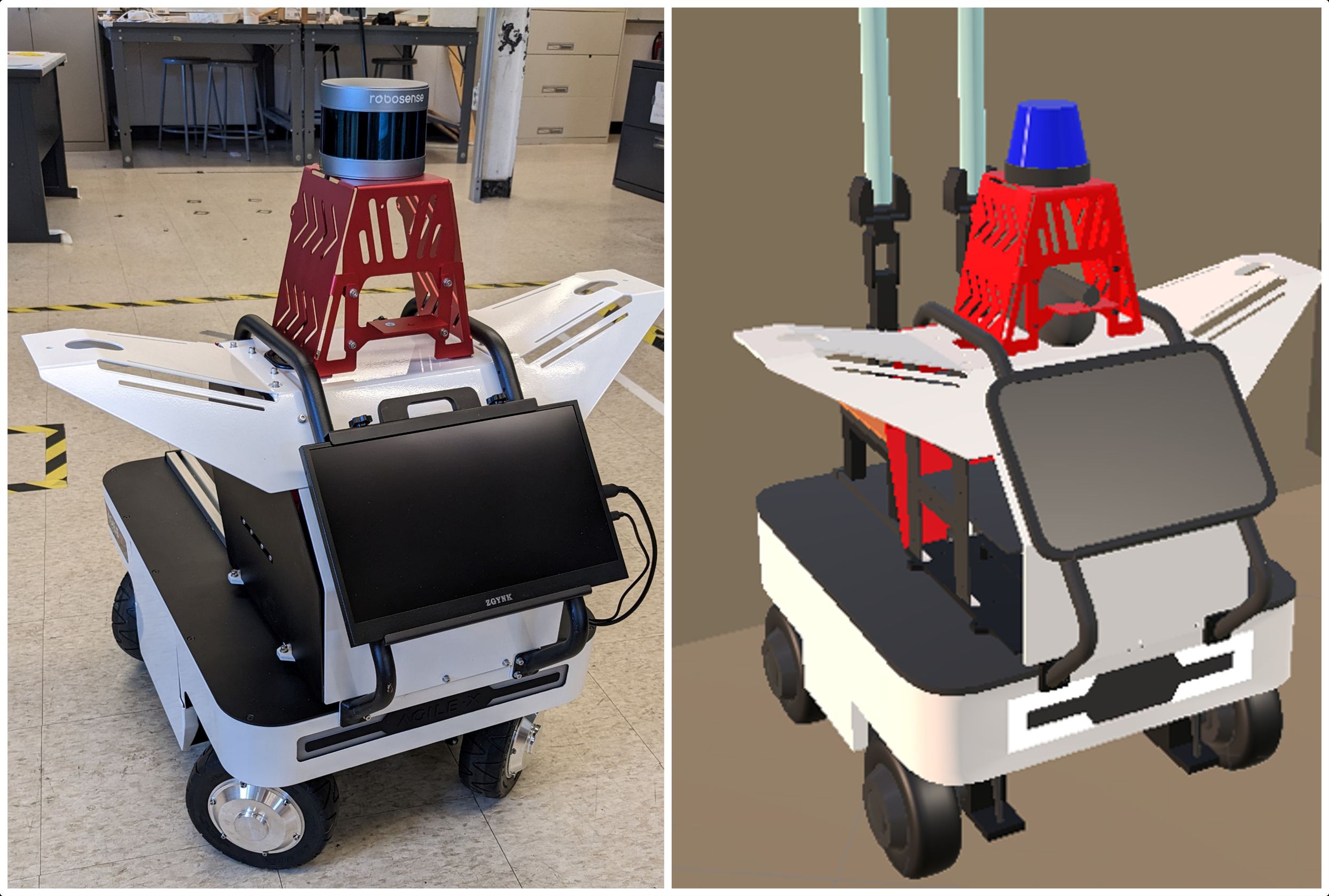}
  \caption{Comparison of the real robot (left) to the  simulated robot (right) used in the experiment.}
  \label{fig:real_vs_sim_RM2}
\end{figure}

\bfhead{Robots.} The robots were designed to simulate the Agile-X Ranger Mini 2.0 (RM2) shown in Figure~\ref{fig:real_vs_sim_RM2}. Similarly to the RM2, the simulated robot could move in all directions. The RM2 has a maximum speed of 1.7 m/s, which is slightly faster than the average walking speed~\cite{fitzpatrick2006another}. Accordingly, we set the maximum speed of the robots to be $130\%$ of the walking speed of the participant in the simulation. The robot’s \textit{approach behavior} was driven by a multi-channel LSTM encoder–decoder model~\cite{xue2018ss, pfeiffer2018data}, which has been successfully applied to pedestrian trajectory prediction. Details of this model can be found in the Appendix. The model takes sequences of prior locations and the relative position of objects as input, while the output is a sequence of predicted future locations. The robots sent raw position data to an off-board computer running the model, and the external computer responded with predictions of the shooter's future positions. 

To create \textit{distractions}, arms with LED lights were added to the robot to wave back and forth. This design is based on previous designs that have been used on evacuation robots~\cite{robinette2014assessment}. A flashing light was also added to the top of the robot, a loud speaker to act as a siren, and a security smoke device, similar to what has been recently sold as a security device~\cite{smokescreenSecuritySmoke}. The smoke device releases puffs of smoke at a frequency of 1.0 Hz, in order to disrupt the vision of the shooter. 

\bfhead{Size of Study.} A pre-experiment power analysis based on pilot studies was conducted to determine the required sample size. Assuming a medium effect size on the number of victims, the analysis indicated that approximately 30 subjects per condition was necessary. A full $2 \times 3$ factorial design (approach type × distraction level) would have required more than 180 participants after accounting for potential dropouts. To make the study feasible while preserving statistical power, two conditions were removed based on observations made from the pilot studies. Pilot studies revealed that participants rarely noticed the robot in the passive-approach, low and medium distraction level conditions. To establish a baseline for comparison, we also included a control condition without a robot. In total, 150 participants were recruited and assigned to conditions using block randomization. Table~\ref{tab:expDesign_distract} depicts the number of subjects in each of the conditions that included a robot.

\begin{table}[h!]
\centering
  \caption{Experimental design for the robot-present conditions, showing the target number of participants per condition. A separate no-robot control group ($n=30$) was included in the study but is not shown here.}
  \label{tab:expDesign_distract}
  \begin{tabular}{|cc|c|c|c|c|}\cline{3-6}
   \multicolumn{2}{c|}{} & \multicolumn{4}{c|}{\textbf{Distraction}}\\ \cline{1-6}
   \vtext{4}{\textbf{Approach}}&\multicolumn{1}{|c|}{} & low & medium & high & total \\ \cline{2-6}
   &\multicolumn{1}{|r|}{passive}   &  0     &   0 & 30    &    30\\ \cline{2-6}
  &\multicolumn{1}{|r|}{aggressive} & 30 & 30 & 30& 90\\ \cline{2-6}
  &\multicolumn{1}{|r|}{total} & 30 & 30 & 60 & 120\\ \cline{1-6}
  \end{tabular}
\end{table}

\bfhead{Ecological Validity.} Admittedly, the data from our study is not based on the actions of real school shooters. Using school shooters as a study subjects is not possible because most school shooters are killed during their attack or are currently serving prison sentences. To maximize ecological validity, we matched our participant demographics to those reported for school shooters. Prior research shows that 96\% of school mass shooters are male with a mean age of 22.8 years ($\pm 8.1$)~\cite{peterson2023violence}. Accordingly, our participants were all male with a mean age of 23.5 years. Prior work has compared the behavior of participants role-playing as school shooters in VR with that of actual school shooters, using measures such as number of shots fired, shot accuracy, and number of victims~\cite{mcclurg2025using}. These studies found the two populations to be statistically equivalent on these key measures. Finally, participants also reported high task seriousness ($M=4.35,\ SD=0.86$) and immersion ($M=3.82,\ SD=0.91$) on a 5-point Likert measure. Taken together, these findings serve as evidence that ecologically valid data can be produced by asking subjects to act as active shooters, although more work will be necessary to fully validate this approach. 

Although it may seem counterintuitive, the behavior of most school shooters is generally straightforward and shows little evidence of being shaped by complex emotions (e.g., fear), crowd dynamics, police presence, or situational factors \cite{JSCOreport, FoxNews_2018, fox2024critical, SandyHookShootingReports}. Rather, school shooters tend to stroll through the environment at a measured (not hurried) pace, engage (rather than avoid) police, and linger in specific areas of the environment. School shooter behavior is typically characterized as deliberative and rational (from the perspective of maximizing harm) and not chaotic, confused, or unpredictable \cite{JSCOreport, FoxNews_2018, fox2024critical, SandyHookShootingReports}. Moreover, historical data investigating the shooter's state of mind during a shooting does not exist, in part because many shooters perish during the attack~\cite{peterson2023violence}. Most data related to shooters captures factors that could serve as a motive or be used to predict who will become a shooter. On the other hand, school shooter behavior during the shooting has been documented as deliberative, calm and focused on generating victims \cite{burrows2022investigate, pirro2013newtown, ausdemore2015eliminating}.

Our study intentionally focuses on shooter behavior and we assume that although unique psychological trauma may motivate the shooter, the actions taken by most shooters in pursuit of their goal to maximize casualties. Essentially, they move through the environment searching for potential victims and attack them. This assumption is supported by numerous police reports and investigations \cite{burrows2022investigate, pirro2013newtown, ausdemore2015eliminating}. As such, our experimental procedure seeks to be face valid -- asking participants to shoot as many NPCs as possible, a task described to each participant three times before the shooting period began and confirmed with post-experiment responses indicating participants took the task seriously.

% -------------------------------------------------------------------
\section{Experimental Results}
Data was collected from October – December 2024 from a total of 172 participants. Data from participants who did not finish the simulation due to physical discomfort (13), emotional discomfort (7), or hardware error (2) were discarded. For reasons described in previous section, the subjects were all male and the average age was 23.5 years. %, which also approximates the target population. %The race distribution of subjects was $43.3\%$ Caucasian, $38.3\%$ Asian, $3.3\%$ Black, and $15.0\%$ other.

\begin{figure}[b!]
  \centering
  \includegraphics[width=\linewidth]{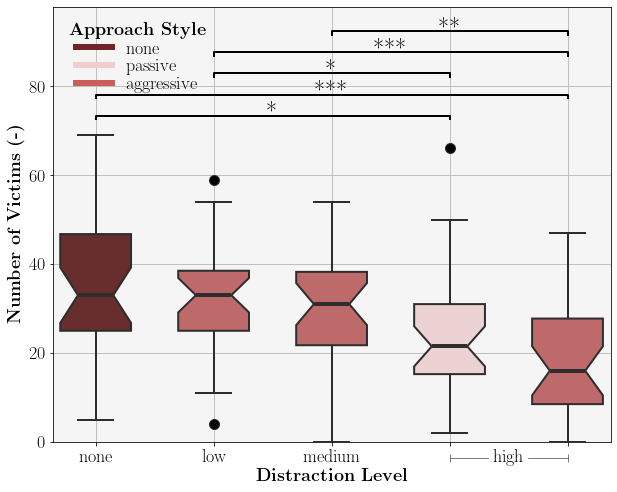}
  \caption{Number of victims ($V$) per robot condition. Boxes represent the interquartile range (IQR), lines indicate medians, and whiskers extend to 1.5 $\times$ IQR. Asterisks indicate significance from Welch's unequal variance $t$-test: *$p < 0.05$, **$p < 0.01$, ***$p < 0.001$.}
  \label{fig:response_vs_victims}
\end{figure}

\bfhead{Hypothesis Testing.} For statistical analysis, we conducted Welch’s unequal-variance $t$-test with a standard significance level of $\alpha=0.05$. Holm-Bonferroni correction~\cite{holm1979simple} was applied to reduce the likelihood of Type I family-wise error, whereas families were determined by grouping strongly correlated dependent variables.\footnote{There were strong correlations between victims and shots fired ($r=0.69$), as well as gaze towards and shots fired at the robot ($r=0.62$).} As this work is \textit{exploratory}, we caution that overly conservative corrections could mask potentially meaningful trends that merit further study. To this end, significance is presented without the Holm-Bonferroni correction unless explicitly stated. For completeness, Table~\ref{tab:posthoc_stats} presents p-values with and without correction.

\begin{figure}[b!]
  \centering
  \includegraphics[width=\linewidth]{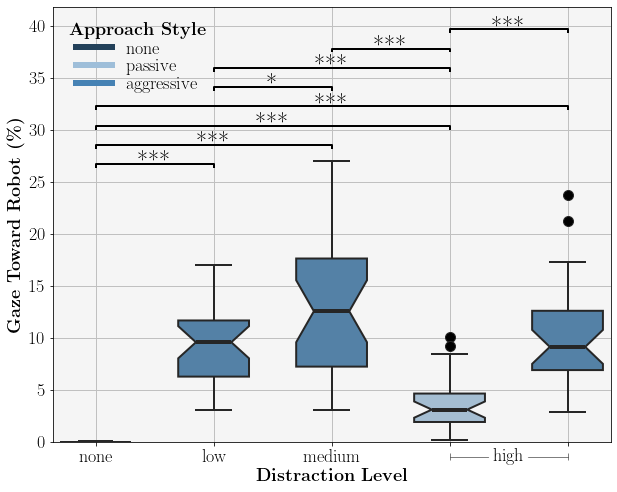}
  \caption{Portion of gaze directed at the robot ($G^R$) per robot condition. Boxes represent the interquartile range (IQR), lines indicate medians, and whiskers extend to 1.5 $\times$ IQR. Asterisks indicate significance from Welch's unequal variance $t$-test: *$p < 0.05$, **$p < 0.01$, ***$p < 0.001$.}
  \label{fig:response_vs_gaze}
\end{figure}

\input{tables/robot_effects}

Our first hypothesis \textbf{(H1)} is partially supported. With an aggressive robot approach, the effect of using light distractors on the number of victims ($V_{low} \rightarrow V_{med}$) was not significant\tval{57.99}{0.863}{.392}. The effect of smoke distractors on the number of victims ($V_{med} \rightarrow V_{high}$) was significant\tval{57.97}{3.146}{.003}. Yet, the effect of both light and smoke distractors on the number of victims ($V_{low} \rightarrow V_{high}$) was significant, $t(57.99)=4.039$, $p<.001$. Our second hypothesis \textbf{(H2)} is rejected. With a high level of robot distraction (smoke, lights, and mere presence), the effect of the increasingly aggressive robot approach on the number of victims ($V_{pass} \rightarrow V_{agg}$) was not significant\tval{57.06}{1.704}{.094}. Our third hypothesis \textbf{(H3)} is partially supported. With an aggressive approach, the effect of light distractors on the shooter's attention to the robot ($G^R_{low} \rightarrow G^R_{med}$) was significant\tval{46.98}{-2.477}{.017}. In contrast, the effect of smoke distractors on the shooter's attention ($G^R_{med} \rightarrow G^R_{high}$) was not significant\tval{54.78}{1.906}{.062}. The effect of lights and smoke on the shooter's attention ($G^R_{low} \rightarrow G^R_{high}$) was not significant\tval{53.95}{-0.467}{0.642}. Our fourth hypothesis \textbf{(H4)} is partially supported. With a high level of robot distraction, the effect of the increasingly aggressive robot approach on the shooter's attention ($G^R_{pass} \rightarrow G^R_{agg}$) was significant, $t(42.46)=-6.524$, $p<.001$. Our fifth hypothesis \textbf{(H5)} is supported. The effect of robot presence on the number of victims ($V_{none} \rightarrow V_{robot}$) was significant, $t(40.38)=2.782$, $p=.008$. More specifically, the best-case condition -- approaching aggressively with smoke and lights -- was significantly greater than having no robot, $t(54.71)=4.359$, $p<.001$. Figures~\ref{fig:response_vs_victims} and~\ref{fig:response_vs_gaze} provide the distributions of the primary dependent variables per robot condition, where significance is annotated. 

The effects of robot intervention are summarized in Table~\ref{tab:posthoc_stats}, where we report both original and adjusted p-values, descriptive statistics, and effect size. With respect to the number of victims ($V$), there is some evidence of a small-to-medium effect when using an aggressive (versus passive) robot approach. There was also some evidence of a small effect in using lights (versus no lights) as distractors. The effect of adding smoke, as well as adding lights and smoke, was large and statistically significant, reducing the number of victims by $35.6\%$ and $41.3\%$, respectively. Finally, the best-case robot condition -- approaching aggressively with smoke and lights -- had a large and significant effect, reducing the number of victims by $46.6\%$ as compared to the control (no robot) condition. With respect to the shooter's attention ($G^R$), there was a large and statistically significant effect when using an aggressive (versus passive) approach, which increased the shooter's gaze towards the robot by $190.5\%$. There was also evidence (significant before adjustment) of a moderate-to-large effect when adding lights as distractors, increasing the shooter's gaze towards the robot by $35.1\%$. Adding smoke as a distractor had a moderately adverse effect, although this observation is more likely due to participants having greater difficulty tracking the robot in a smoky environment, rather than because they attended to the robot less.

\bfhead{Subjective Responses. }After completing the experiment, participants were asked to rate the importance, helpfulness, and perceived danger of the study (Table~\ref{tab:subjectiveResponses}). The majority rated the study as very to extremely important (86\%) and helpful (75\%), while most reported that the experience was not at all or only slightly dangerous (75\%). These findings suggest that participants engaged seriously with the task and viewed the research as valuable, supporting the credibility of the data.

\input{tables/subjective_responses}

\section{Discussion} 
Our findings demonstrate the potential of robot-based school shooting interventions and also shed light on why certain conditions are more effective. The sharp reduction in victim counts under the aggressive, high-distraction condition suggests that multi-modal sensory disruption -- particularly smoke -- overloaded participants’ perception and hindered their ability to target. The gaze data revealed that smoke made the robot harder to track, which may have further diverted attention away from victims. These results suggest that the most effective interventions may not rely on continuous shooter engagement but rather on forcing momentary attention shifts that interrupt attack sequences. This insight has design implications for future response robots. First, multi-modal distractors should be prioritized over single-channel cues such as lights or sirens. Second, an aggressive approach strategy appears to enhance distraction by forcing visual confrontation, though future systems may need to balance aggressiveness against safety for bystanders. Third, our results motivate real-time adaptation: a robot might escalate from passive to aggressive behaviors or vary distractor intensity based on the shooter’s current state (e.g., proximity to victims, orientation toward exits).

\bfhead{Ethical Considerations. }Our research raises a variety of ethical issues. The risk of over-reliance, false positives and negatives, and intentional misuse (e.g., triggering the robot for fun) must be weighed against the system's potential for saving lives. Moreover, communication and rescue efforts could be impacted if the robot malfunctions, gives students a false sense of escapability, or obstructs first responders. Because of the high-stakes nature of this work, additional research will be needed before deploying and relying on a robot for this purpose. One motivation for our research is therefore to assess the potential risks and benefits and to begin the discussion about the broader impacts of this application. While new technologies often face initial skepticism \cite{fischhoff1979weighing, banta1979weighing}, their benefits can outweigh carefully managed risks if potential stakeholders are engaged and the system is carefully and systematically evaluated at each step of the development process. The work presented in this paper is one step in this process.

\bfhead{Participant Safety. }In addition to societal and deployment risks, we considered the potential impact of this study on the participants themselves. We do not believe that participants were emotionally harmed by participating in this research. Post-experiment surveys indicated that over 75\% of participants felt the study was either not-at-all or only slightly dangerous (see Table~\ref{tab:subjectiveResponses}). Participants were told that they could quit at any point if they felt physical and/or emotional discomfort and would still receive full payment; only 7 out of 172 participants chose to withdraw for this reason. To minimize risk further, we provided all participants with the contact information for university mental health services should they experience distress after leaving the site of the experiment. The experiment was intentionally designed to be face valid, avoiding deception or surprise, and was approved by the Institutional Review Board (IRB).

\bfhead{Limitations. }Several limitations should be considered when interpreting these results. First, this study necessarily relied on participants acting as shooters rather than real offenders, as direct access to actual school shooters is not feasible. However, prior work and our own participant surveys indicate that behavior in this paradigm is statistically equivalent to historical data on key measures such as shot rate, victim rate, and accuracy~\cite{mcclurg2025using}. Second, the experiment used scripted NPC behaviors based on the widely recommended ``Run. Hide. Fight'' protocol, which reflects observed victim behavior but may not capture the full variability and social dynamics of real incidents~\cite{ausdemore2015eliminating, pirro2013newtown}. Third, robot behaviors were limited to a small set of distraction conditions and did not account for potential adversarial adaptation or multiple coordinated attackers. Instead, this study intentionally isolated the effect of several key distractors in the most common shooting paradigm: single-shooter incidents~\cite{peterson2023violence}. Finally, the simulated robots were idealized and did not include sensor noise, actuation error, or real-world delays. Nonetheless, the distraction modalities themselves -- loud noises, bright lights, and smoke -- are physical phenomena likely to have similar effects in practice. 

% -------------------------------------------------------------------
\section{Conclusion} 
This paper explores the application of robots to delay and/or distract school shooters. Our human subject experiment, which asks participants to act as a school shooter in virtual reality, provides evidence that a robot can be used to reduce the number of victims in a shooting event. One specific robot condition -- aggressively approaching the shooter with both lights and smoke -- significantly reduced the number of victims by $46.6\%$, compared to having no robot. Beyond the empirical results, this work contributes a framework for safe, reproducible study of high-risk and adversarial human-robot interaction scenarios. By combining immersive virtual reality with well-defined robot behaviors, our approach enables systematic and ethically viable testing of robot strategies under conditions that would be infeasible or unsafe to reproduce in the real world. 

Future work will increase the fidelity and scope of the simulation environment rather than attempting unsafe real-world validation. We plan to incorporate more realistic robot models with simulated sensor noise to better approximate real-world performance. Additional experiments will generate training data for empirical models of victims and first responders. Failure modes should also be studied, such as quantifying side effects of distraction (e.g., whether smoke slows victim evacuation) to ensure that benefits outweigh unintended harm. The experiment could also be extended to explore multiple and/or adaptive shooters, where a shooter could learn to ignore the robot, or simply destroy it. Exploring such scenarios could inform stochastic or adversarial planning methods that keep robots unpredictable. Finally, adding outcome measures beyond victim counts could further refine what constitutes a successful intervention. Together, these efforts will produce more generalizable design guidelines and advance the responsible development of robots for school safety. 

% -------------------------------------------------------------------
\begin{acks}
This material is based upon work supported by the National Science Foundation under Grant No.~IIS-2045146. Any opinions, findings, and conclusions or recommendations expressed in this material are those of the author(s) and do not necessarily reflect the views of the National Science Foundation.
\end{acks}

% -------------------------------------------------------------------
\bibliographystyle{plainnat}
\bibliography{references}

\end{document}

\input{appendix}

%% file: figures/procedure.tex
\begin{figure}[!b]
\centering
\begin{tikzpicture}[
    node distance=0.9cm and 1.6cm,
    every node/.style={font=\small, align=center},
    box/.style={draw, rounded corners, fill=gray!10, 
                inner sep=3pt, minimum width=3.2cm, minimum height=0.8cm},
    arrow/.style={-{Latex[length=2mm]}, thick}
]

% Left column (early steps)
\node[box] (consent) {Consent form};
\node[box, below=of consent] (vr) {VR equipment};
\node[box, below=of vr] (train) {Practice moving \\\& shooting};
\node[box, below=of train] (walk) {Walkthrough school\\points of interest};

% Right column (later steps)
\node[box, right=of consent] (gun) {Handgun issued};
\node[box, below=of gun] (npcs) {NPCs arrive};
\node[box, below=of npcs] (shooting) {Shooting period\\(5 min duration)};
\node[box, below=of shooting] (survey) {Post-experiment survey};

% Arrows left column
\draw[arrow] (consent) -- (vr);
\draw[arrow] (vr) -- (train);
\draw[arrow] (train) -- (walk);

% Arrow from left to right
\draw[arrow] (walk.east) -- ++(0.8,0) |- (gun.west);

% Arrows right column
\draw[arrow] (gun) -- (npcs);
\draw[arrow] (npcs) -- (shooting);
\draw[arrow] (shooting) -- (survey);
\end{tikzpicture}
\caption{Each participant followed the procedure above upon arrival. Once the experiment was complete, they were paid. Participants could quit at any time if they felt physical and/or mental discomfort and they would still be paid.}
\label{fig:procedure}
\end{figure}

%% file: tables/robot_effects.tex
\begin{table*}[!t]
    \centering
    \caption{Summary of the reported effects of robot intervention. Post-hoc adjustment of p-values is determined by Holm-Bonferroni~\cite{holm1979simple} method. Checkmarks indicate respective p-values are significant ($p<0.05$).}
    \label{tab:posthoc_stats}
    \renewcommand{\arraystretch}{1.5}
    \begin{tabular}{|c|c|cc|cc|c|c|c|c|}
    \hline
        \textbf{} & \textbf{Effect} & \multicolumn{2}{c|}{\textbf{p-value}} & \multicolumn{2}{c|}{\textbf{p-value}} & \textbf{Mean} & \textbf{Mean} & \textbf{Mean}  & \textbf{Effect}   \\[-0.1cm] 
        \textbf{Hypothesis} & \textbf{$\boldsymbol{A} \rightarrow\boldsymbol{B}$} & \multicolumn{2}{c|}{\textbf{(original)}} & \multicolumn{2}{c|}{\textbf{(adjusted)}} & \textbf{(Condition A)} & \textbf{(Condition B)} & \textbf{Difference ($\%$)} & \textbf{Size,  $\lvert d\rvert$}   \\ \hline 
        \multirow{3}{*}{$1$} & $V_{low} \rightarrow V_{med}$ & $0.392$ &  & $0.392$ &   & $32.10 \pm 12.57$ & $29.23 \pm 12.73$ & $-8.9\%$ & $0.22$\\ \cline{2-10}
        & $V_{med} \rightarrow V_{high}$ & $0.003$ & \cm  & $0.021$ & \cm   & $29.23 \pm 12.73$ & $18.83 \pm 12.45$ & $-35.6\%$ & $0.84$\\ \cline{2-10}
        & $V_{low} \rightarrow V_{high}$ & $<.001$ & \cm  & $0.001$ & \cm &   $32.10 \pm 12.57$ & $18.83 \pm 12.45$ & $-41.3\%$ & $1.06$\\ \hline
        $2$ & $V_{pass} \rightarrow V_{agg}$ & $0.094$ &  & $0.469$ &  &   $24.80 \pm 14.17$ & $18.83 \pm 12.45$ & $-24.1\%$ &  $0.45$ \\ \hline
        \multirow{3}{*}{$3$} & $G^R_{low} \rightarrow G^R_{med}$ & $0.017$ & \cm & $0.118$ &   & $9.55 \pm 3.69$ & $12.90 \pm 6.26$ & $+35.1\%$ & $0.65$\\ \cline{2-10}
        & $G^R_{med} \rightarrow G^R_{high}$ & $0.062$ &  & $0.371$ & & $12.90 \pm 6.26$ & $10.08 \pm 4.89$ & $-21.9\%$ & $0.50$\\ \cline{2-10}
        & $G^R_{low} \rightarrow G^R_{high}$ & $0.642$ &  & $1.285$ & & $9.55 \pm 3.69$ & $10.08 \pm 4.89$ & $+5.5\%$ & $0.10$\\ \hline
        \multirow{1}{*}{$4$} & $G^R_{pass} \rightarrow G^R_{agg}$ & $<.001$ & \cm & $<.001$ & \cm  & $3.47 \pm 2.43$ & $10.08\pm 4.89$ & $+190.5\%$ & $1.71$\\ \hline
        \multirow{2}{*}{$5$}& $V_{none} \rightarrow V_{robot}$ & $0.008$ & \cm & $0.057$ &  & $35.23 \pm 15.99$ & $26.24\pm 13.93$ & $-25.5\%$ & $0.60$\\ \cline{2-10}
        & $V_{none} \rightarrow V_{best}$ & $<.001$ & \cm & $0.001$ & \cm  & $35.23 \pm 15.99$ & $18.83\pm 12.45$ & $-46.6\%$ & $1.14$\\ \hline
    \end{tabular}
    \renewcommand{\arraystretch}{1}
\end{table*}

%% file: tables/subjective_responses.tex
\begin{table}[h!]
\small
\centering
  \caption{Anonymous responses from surveyed participants (n=102) after the study.}
  \label{tab:subjectiveResponses}
  \setlength{\tabcolsep}{4pt}
  \begin{tabular}{|l|c|c|c|c|c|}\hline
    \multicolumn{1}{|c|}{\textbf{Quality}} & \textbf{Not at all} & \textbf{Slightly} &  \textbf{Moderately} & \textbf{Very} & \textbf{Extremely}\\ \hline
    Important & 0 & 4 & 10 & 43 & 45\\ \hline
    Helpful & 1 & 3 & 22 & 36 & 40\\ \hline
    Dangerous & 38 & 39 & 21 & 1 & 3\\ \hline
  \end{tabular}
\end{table}

%% file: appendix.tex
\newpage
\clearpage
\appendix

\section{Shooter Prediction Model}
\begin{figure*}[h]
  \centering
  \includegraphics[width=\linewidth]{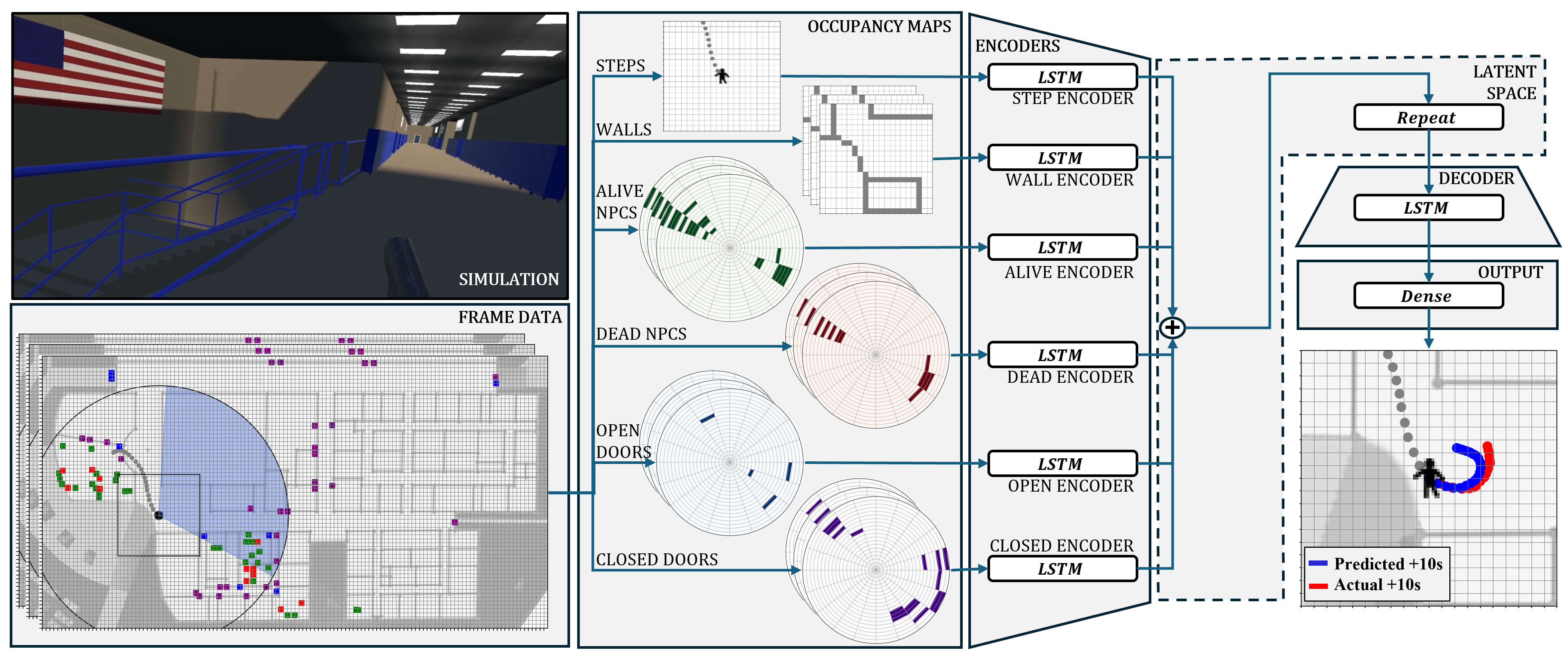}
  \caption{An overview of our shooter trajectory model. Previous trajectory and object maps are encoded by an LSTM layer. The final states of the encoder are decoded, concatenated with the context vector, and passed through a dense layer for trajectory prediction.}
  \label{fig:sed}
\end{figure*}

The predictive model for shooter trajectory is depicted in Figure~\ref{fig:sed}. In our implementation, each object occupancy map (except walls) used a polar grid with 20 angular steps and 20 radial steps with maximum radius was 100 feet. The wall occupancy map used a Cartesian grid with 20 steps in each direction. Thus, the input for each object encoder was $(n_{prev}, 400)$. Each object encoder contained 100 LSTM cells, and thus, resulted in final cell states and hidden states with size $(1, 100)$. After concatenation, the final cell states and final hidden states had size $(1, 600)$. The final cell states were used to initialize the LSTM decoder, while the final hidden states were repeated over each prediction step. The decoder contained 600 LSTM cells such that the size of the returned sequence was $(n_{pred}, 600)$. Finally, this output was passed through a fully-connected, dense layer to predict a sequence of steps. 

